\documentclass{article}

\usepackage{microtype}
\usepackage{graphicx}
\usepackage{subfigure}
\usepackage{booktabs} 

\usepackage{arydshln}
\usepackage{multirow} 
\usepackage{multicol}
\usepackage{tabularx}

\usepackage{hyperref}

\usepackage[accepted]{icml2024}

\usepackage{amsmath}
\usepackage{amssymb}
\usepackage{mathtools}
\usepackage{amsthm}

\usepackage[capitalize,noabbrev]{cleveref}

\theoremstyle{plain}

\theoremstyle{definition}

\theoremstyle{remark}

\usepackage[textsize=tiny]{todonotes}

\icmltitlerunning{Textual Localization: Decomposing Multi-concept Images for Subject-Driven Text-to-Image Generation}

\begin{document}

\twocolumn[
\icmltitle{Textual Localization: Decomposing Multi-concept Images for Subject-Driven Text-to-Image Generation}




\begin{icmlauthorlist}
\icmlauthor{Junjie Shentu}{yyy}
\icmlauthor{Matthew Watson}{yyy}
\icmlauthor{Noura Al Moubayed}{yyy}
\end{icmlauthorlist}

\icmlaffiliation{yyy}{Department of Computer Science, Durham University, Durham, UK}

\icmlcorrespondingauthor{Noura Al Moubayed}{noura.al-moubayed@durham.ac.uk}

\icmlkeywords{Machine Learning, ICML}

\vskip 0.3in
]



\printAffiliationsAndNotice{}  

\begin{abstract}
Subject-driven text-to-image diffusion models empower users to tailor the model to new concepts absent in the pre-training dataset using a few sample images. However, prevalent subject-driven models primarily rely on single-concept input images, facing challenges in specifying the target concept when dealing with multi-concept input images. To this end, we introduce a textual localized text-to-image model (\textit{Texual Localization}) to handle multi-concept input images. During fine-tuning, our method incorporates a novel cross-attention guidance to decompose multiple concepts, establishing distinct connections between the visual representation of the target concept and the identifier token in the text prompt. Experimental results reveal that our method outperforms or performs comparably to the baseline models in terms of image fidelity and image-text alignment on multi-concept input images. In comparison to \textit{Custom Diffusion}, our method with hard guidance achieves CLIP-I scores that are 7.04\%, 8.13\% higher and CLIP-T scores that are 2.22\%, 5.85\% higher in single-concept and multi-concept generation, respectively. Notably, our method generates cross-attention maps consistent with the target concept in the generated images, a capability absent in existing models.
\end{abstract}

\section{Introduction}
\label{introduction}

Diffusion models have demonstrated unprecedented capabilities in generating high-quality and diverse images, effectively addressing the mode collapse problem encountered by Generative Adversarial Networks (GANs) \cite{ho2020denoising, dhariwal2021diffusion}. By leveraging cross-attention layers within the UNet architecture \cite{ronneberger2015u}, diffusion models can be conditioned on information in different modalities. Text-to-image diffusion models, a specific subclass of conditional diffusion models, exemplify remarkable proficiency in producing images semantically aligned with natural language prompts \cite{nichol2021glide, rombach2022high, ramesh2022hierarchical, gu2022vector}.

Despite the strong image-text connections established by text-to-image models, introducing new concepts not present in the pre-training datasets remains challenging \cite{gal2022image}. In response, studies aimed at ``customizing" text-to-image models for generalization to newly introduced concepts propose fine-tuning pre-trained models with a few (typically 3 to 5) images of the target object, defining them as subject-driven text-to-image models \cite{gal2022image, ruiz2023dreambooth, kumari2023multi, li2023blip, jia2023taming, gal2023encoder, arar2023domain, ruiz2023hyperdreambooth, ma2023subject}. However, prevalent subject-driven models are designed to learn from images containing a single new concept (termed single-concept images), imposing high requirements for data preparation and demanding a prolonged fine-tuning process when introducing multiple concepts \cite{kumari2023multi}. Conversely, the feasibility of applying images containing multiple concepts (termed multi-concept images) for fine-tuning has not been thoroughly explored.

\begin{figure}[htb]
\vskip 0.1in
\begin{center}
\centerline{\includegraphics[width=\columnwidth]{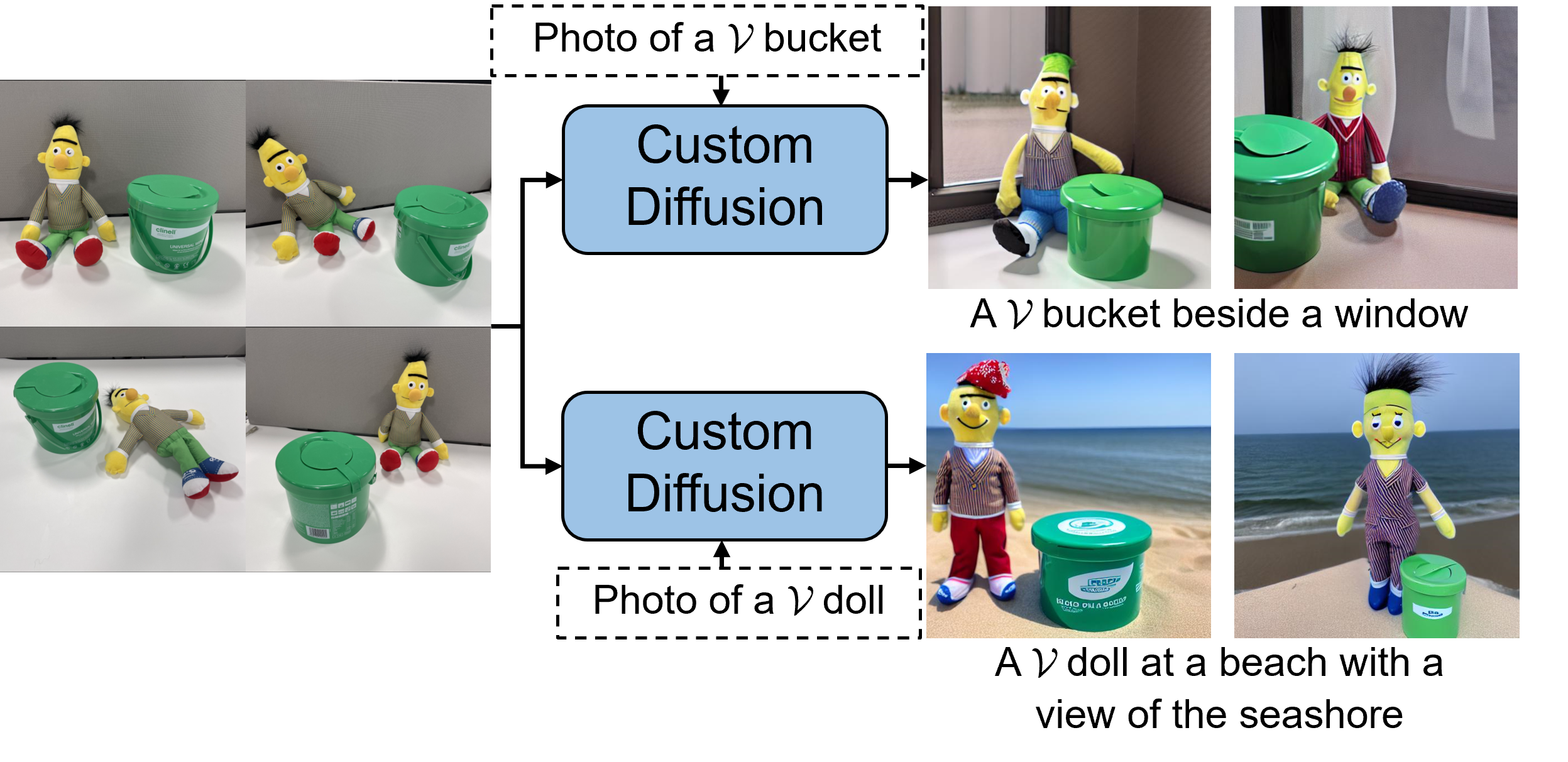}}
\caption{Failure cases in single-concept generation by Custom Diffusion when fine-tuning on multi-concept inputs}
\label{fig1}
\end{center}
\vskip -0.1in
\end{figure}

To this end, we evaluate the state-of-the-art model \textit{Custom Diffusion} \cite{kumari2023multi} on multi-concept images. The test involves utilizing the text prompt ``\textit{Photo of a} $\mathcal{V}$ [$class$]", where $\mathcal{V}$ serves as a unique identifier token representing the target concept, and [$class$] denotes the class name of the concept. We find that the model tends to generate all concepts present in the input images, disregarding the specified target concept in the text prompt. Example instances of these failure cases are showcased in \cref{fig1}.

To address the identified deficiencies in existing subject-driven models, we propose a novel approach named \textit{Textual Localization}. This method aims to decompose input images and achieve precise customization of the target concepts, especially when confronted with multi-concept images.  During model fine-tuning, we introduce a cross-attention guidance mechanism that incorporates a new cross-attention loss $L_{attn}$, which is designed to pinpoint the target concept region in the input image and establish a distinct connection between the visual representation of the target concept and the identifier token $\mathcal{V}$. Our method encompasses two distinct strategies of cross-attention guidance: hard guidance and soft guidance, both of which eliminate the model's attention on non-target concepts but apply different ways of attention activation in the target region of the input images. The cross-attention guidance manipulates the cross-attention maps between the input images and identifier token $\mathcal{V}$, thereby influencing the model's attention. Additionally, we compile a dataset comprising both multi-concept images and single-concept images for model training and evaluation. Experimental results indicate that our method either outperforms or matches baseline models in both single-concept and multi-concept generation when taking multi-concept images as input. Moreover, our approach explicitly showcases the connection between the visual representation of the target concept and identifier token $\mathcal{V}$ through the cross-attention maps.

\section{Related Work}
\subsection{Text-to-image diffusion model}
Diffusion model, functioning as a likelihood-based model, attains state-of-the-art performance in generating high-quality images, surpassing other generative models \cite{ho2020denoising, dhariwal2021diffusion}. Additionally, the inclusion of cross-attention layers equips the diffusion model with the capability to incorporate conditioning information in diverse modalities \cite{rombach2022high}, with natural language being one of the predominant sources of conditioning information.  \citet{nichol2021glide} and \citet{saharia2022photorealistic} apply a text-to-image generation model in the pixel space with classifier-free guidance \cite{ho2022classifier}. They utilize a transformer \cite{vaswani2017attention} and a pre-trained large language model \cite{ramesh2021zero} as text encoders, respectively. Besides, \citet{rombach2022high} train a diffusion model in the latent space by using a Variational Autoencoder (VAE) \cite{kingma2013auto} to project images into latent space, using a pre-trained text encoder from Contrastive Language-Image Pre-training (CLIP) \cite{radford2021learning} to process the text prompt. \citet{ramesh2022hierarchical} propose a multimodal latent space by training a prior model to generate the CLIP image embedding of the input text prompt, and generating images conditioned on the image embedding.

\subsection{Subject-driven text-to-image generation}
To address the challenge of generating novel concepts absent in the pre-training dataset, subject-driven generation is devised to customize the text-to-image generation model using a limited set of sample images \cite{ruiz2023dreambooth}. In subject-driven generation, rare-occurring identifier tokens from the vocabulary are inserted into the text prompt to establish a connection with the input images during the fine-tuning process. Various training targets are explored, including solely the text encoder \cite{gal2022image}, the text encoder along with cross-attention layers in the UNet \cite{kumari2023multi, shi2023instantbooth}, and the text encoder together with the entire UNet \cite{ruiz2023dreambooth}. Additionally, \citet{li2023blip} and \citet{jia2023taming} incorporate an image encoder to obtain more accurate and robust embeddings of the input images, replacing the identifier tokens. In a bid to further optimize the fine-tuning process and reduce the number of training parameters, \citet{arar2023domain} and \citet{ruiz2023hyperdreambooth} propose applying Low-Rank adaptation (LoRA) \cite{hu2021lora} for expedited personalization. However, the majority of the aforementioned models predominantly focus on learning from single-concept images, while our approach excels in decomposing input images and facilitating precise learning from multi-concept images.

\subsection{Cross-attention in text-to-image image generation}
Diffusion models harness the cross-attention layers \cite{vaswani2017attention} embedded in the underlying UNet backbone to integrate conditioning information from text prompts into the generated images \cite{rombach2022high}. These cross-attention layers amalgamate information from both images and text, producing cross-attention maps that represent the probability distribution over text tokens for each image patch in the latent space \cite{tang2022daam, chefer2023attend}. Guiding these cross-attention maps during inference empowers the pre-trained diffusion model to generate images with superior semantic alignment to the provided text prompts \cite{feng2022training, chefer2023attend, wang2023compositional, phung2023grounded}, achieve image editing \cite{hertz2022prompt}, and provide positional control over the contents in the generated images \cite{ma2023directed, he2023localized, chen2023training, phung2023grounded}. Besides, cross-attention guidance is also applied during training to help achieve zero-shot personalized image generation, although the generation quality is inferior compared to the models fine-tuned on input images \cite{ma2023subject}. \citet{xiao2023fastcomposer} utilize cross-attention guidance to address the identity blending problem and enable multi-subject generation. Notably, their model's performance is demonstrated on a human face dataset, with its general subject performance remaining undisclosed.

\section{Textual Localized Diffusion Model}


\subsection{Preliminaries}
\label{sec3.1}
In this study, we adopt Stable Diffusion (SD) as the foundational model, built upon the Latent Diffusion Model (LDM) \cite{rombach2022high}. For an input image $x \in \mathbb{R}^{H\times W\times 3}$, SD initially projects $x$ into a latent representation $z \in \mathbb{R}^{h\times w\times c}$ by employing the encoder $\mathcal{E}$ of a VAE \cite{kingma2013auto}, where $c$ denotes the latent feature dimension. The downsampling follows a factor $f=H/h=W/w$, determining the downsampling scale. The diffusion process is subsequently executed on the latent representation by introducing noise into $z$, forming a fixed-length Markov Chain denoted as $z_1 \dots z_T$, where $T$ signifies the length of the chain. SD trains the UNet to learn the reverse process of the Markov Chain, predicting a denoised variant of the input $z_t$ given the timestep $t \in [1, T]$. In the context of text-to-image generation, the conditioning information $y$ from the text prompts is projected into an intermediate representation $\tau_\theta(y)$, where $\tau_\theta$ is a pre-trained CLIP text encoder. The training objective of the text-to-image diffusion model can be expressed as:
\begin{equation}
\label{1}
    L_{LDM} = \mathbb{E}_{\mathcal{E}(x),y,t} \left [ \left \| \varepsilon - \varepsilon_{\theta} \left ( z_t,t,\tau_\theta(y) \right )  \right \|^2_2 \right ] 
\end{equation}
where $\varepsilon$ and $\varepsilon_\theta$ represent the standard Gaussian noise ($\varepsilon \sim \mathcal{N}(0,1)$) and predicted noise residue, respectively. Specifically, the intermediate representation $\tau_\theta(y)$ is linked to the intermediate layers of the UNet through cross-attention layers using the following mapping:
\begin{equation}
\label{2}
    Attention(Q,K,V)=softmax \left ( \frac{QK^T}{\sqrt{d}} \cdot V \right ) 
\end{equation}
\begin{equation}
\label{3}
    Q=W_{Q}^{(i)} \cdot \varphi_i(z_t), K=W_{K}^{(i)} \cdot \tau_\theta(y), V=W_{V}^{(i)} \cdot \tau_\theta(y)
\end{equation}
where $d$ signifies the output dimension of the query ($Q$) and key ($K$) features. $W_{Q}^{(i)}$, $W_{K}^{(i)}$, and $W_{V}^{(i)}$ are learnable projection matrices in cross-attention layer $i$, $\varphi_i(z_t)$ is a flattened intermediate representation of the noisy latent $z_t$. The cross-attention map at layer $i$ is given by:
\begin{equation}
\label{4}
    Attn^{(i)}=softmax \left ( \frac{Q^{(i)}K^{{(i)}^T}}{\sqrt{d}} \right ) 
\end{equation}

\subsection{Pipeline of Textual Localization}
\label{sec3.2}
Subject-driven text-to-image models establish a connection between the new concept from the input images and the identifier token $\mathcal{V}$. During the fine-tuning process, the text embedding of $\mathcal{V}$ is refined to represent the target concept through the cross-attention layers, and the model learns a connection between the text embedding of $\mathcal{V}$ with the visual representation of the target concept in pixel space \cite{ruiz2023dreambooth, kumari2023multi}. However, when presented with multi-concept images, this connection becomes ambiguous, as depicted in \cref{fig1}. To address this ambiguity, we enhance the model by incorporating additional cross-attention guidance during the fine-tuning process. Our proposed method is denoted as the textual localized text-to-image model, or \textit{Textual Localization}.

\begin{figure}[htb]
\vskip 0.1in
\begin{center}
\centerline{\includegraphics[width=\columnwidth]{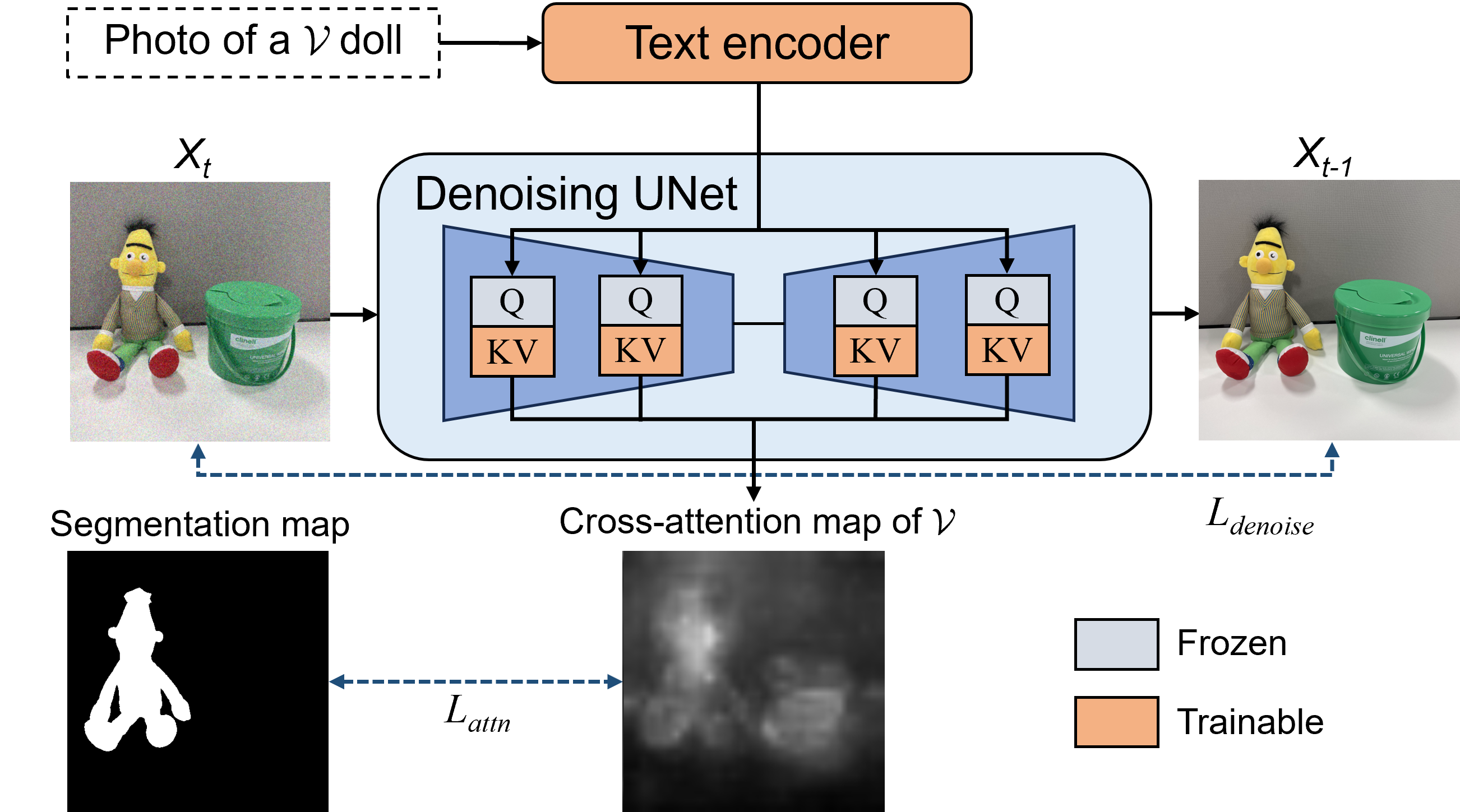}}
\caption{Illustration of a single step of the fine-tuning process of \textit{Textual Localization}}
\label{fig2}
\end{center}
\vskip -0.1in
\end{figure}

A single fine-tuning step of \textit{Textual Localization} is illustrated in \cref{fig2}. Following \textit{Custom Diffusion}, we only optimize the text encoder as well as the $W_K$ and $W_V$ matrices in the cross-attention layers in the UNet. The learning objective of the \textit{Textual Localization} involves minimizing a loss function comprised of a denoising loss $L_{denoise}$ and an attention loss $L_{attn}$. The denoising loss encompasses the original training objective of LDM, as given by \cref{1}, along with a class-specific prior preservation loss, expressed as:
\begin{equation}
\label{5}
    L_{prior} = \mathbb{E}_{\mathcal{E}(x_{pr}),y_{pr},t} \left [ \left \| \varepsilon - \varepsilon_{\theta} \left ( z_{{pr}_t},t,\tau_\theta(y_{pr}) \right )  \right \|^2_2 \right ] 
\end{equation}
where $x_{pr}$ is the sample generated by the pre-trained text-to-image model under the text prompt $y_{pr}$ that solely specifies the class name of the target concept. The inclusion of the class-specific prior preservation loss serves to maintain output diversity and prevent language drift \citep{ruiz2023dreambooth}. The cross-attention loss $L_{attn}$ is formulated to bias the model's attention, establishing a clear connection between the identifier token $\mathcal{V}$ and the target concept. The intricacies of cross-attention guidance are elucidated in \cref{sec3.3}. Consequently, the overarching training objective is to minimize:
\begin{equation}
\label{8}
    L = L_{LDM} + \lambda L_{prior} + \delta L_{attn}
\end{equation}
where $\lambda$ and $\delta$ are two scaling coefficients.

\subsection{Cross-attention Guidance}
\label{sec3.3}
The positional information of the target concept in the input images is derived from the segmentation maps, generated by a pre-trained segmentation model. Through the utilization of these segmentation maps, we introduce two distinct strategies for cross-attention guidance: hard guidance and soft guidance. 

\textbf{Hard guidance.} In the case of hard guidance, the cross-attention map of the identifier token  $Attn_{\mathcal{V}}$ is optimized to align with the segmentation map $Seg \in \mathbb{R}^{H'\times W'}$. The attention loss $L_{attn}$ is computed as the mean square error (MSE) between $Attn_{\mathcal{V}}$ and $Seg$. Thus, $L_{attn}$ can be formulated as:
\begin{equation}
\label{6}
    L_{attn} = \frac{1}{H'W'} {\textstyle \sum_{i=1}^{H'}} {\textstyle \sum_{j=1}^{W'}} \left ( Seg(i,j) - Attn_{\mathcal{V}}(i,j) \right )^2
\end{equation}
Note that $L_{attn}$ is calculated in the pixel space. Given that the cross-attention maps $Attn^{(i)}$ extracted from various layers possess distinct resolutions determined by their positions in the UNet, all cross-attention maps of the identifier token $Attn_{\mathcal{V}}^{(i)}$ are up-scaled to $H' \times W'$ and subsequently averaged to obtain $Attn_{\mathcal{V}}$. 

\textbf{Soft guidance.} The objective of soft guidance is to eliminate the model's attention on regions outside of the target concept in the input images, without influencing attention within the target concept region. The attention loss $L_{attn}$ is formulated as the element-wise product of a binary matrix ($B_{Inv}$), representing the inverse segmentation map, and the MSE between $Attn_{\mathcal{V}}$ and $Seg$, which is given by:

\begin{equation}
\label{7}
 \begin{split}
       L_{attn} = &\frac{1}{H'W'} {\textstyle \sum_{i=1}^{H'}} {\textstyle \sum_{j=1}^{W'}} \\
       &\left [ \left ( Seg(i,j) - Attn_{\mathcal{V}}(i,j) \right )^2 \cdot B_{Inv}(i,j) \right ]
 \end{split}
\end{equation}

\begin{equation}
\label{9}
       B_{Inv}(i,j) = \begin{cases}
                            1 & \text{ if } Seg(i,j) = 0 \\
                            0 & \text{ if } Seg(i,j) > 0
                        \end{cases}
\end{equation}

While both hard guidance and soft guidance effectively reduce the model's attention on non-target concepts, they diverge in their treatment of the target concept region in the input images. Notably, hard guidance influences the model to activate attention in the target region, producing cross-attention maps that align with the segmentation map. In contrast, soft guidance does not alter attention activation in the region of the target concept.

\section{Experiments and Results}
\label{experiments}

\begin{figure*}[htb!]
\vskip 0.1in
\begin{center}
\centerline{\includegraphics[width=1.9\columnwidth]{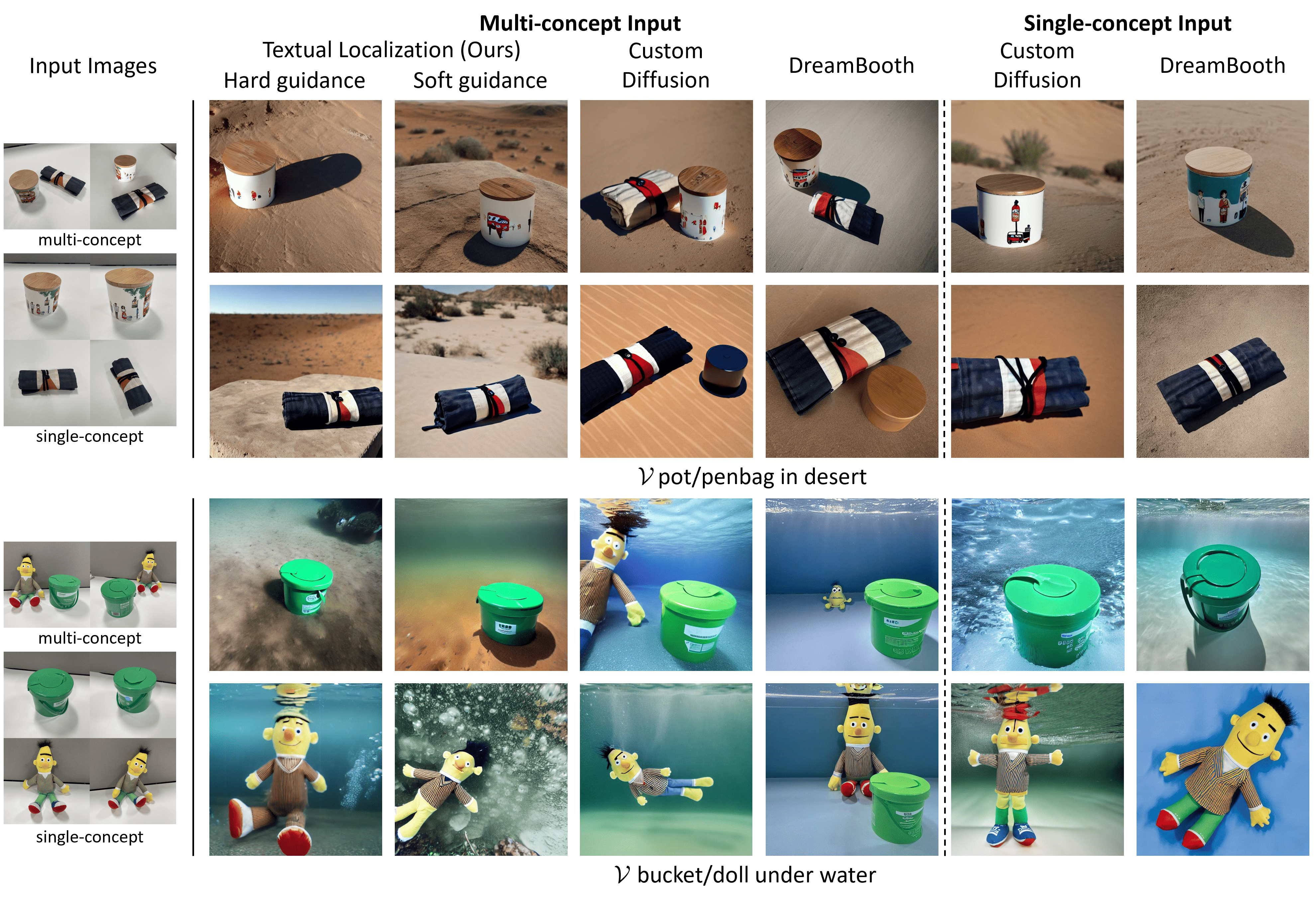}}
\caption{Qualitative comparison in single-concept generation}
\label{single_subject_result}
\end{center}
\vskip -0.1in
\end{figure*}

\begin{figure*}[htb!]
\vskip 0.1in
\begin{center}
\centerline{\includegraphics[width=1.9\columnwidth]{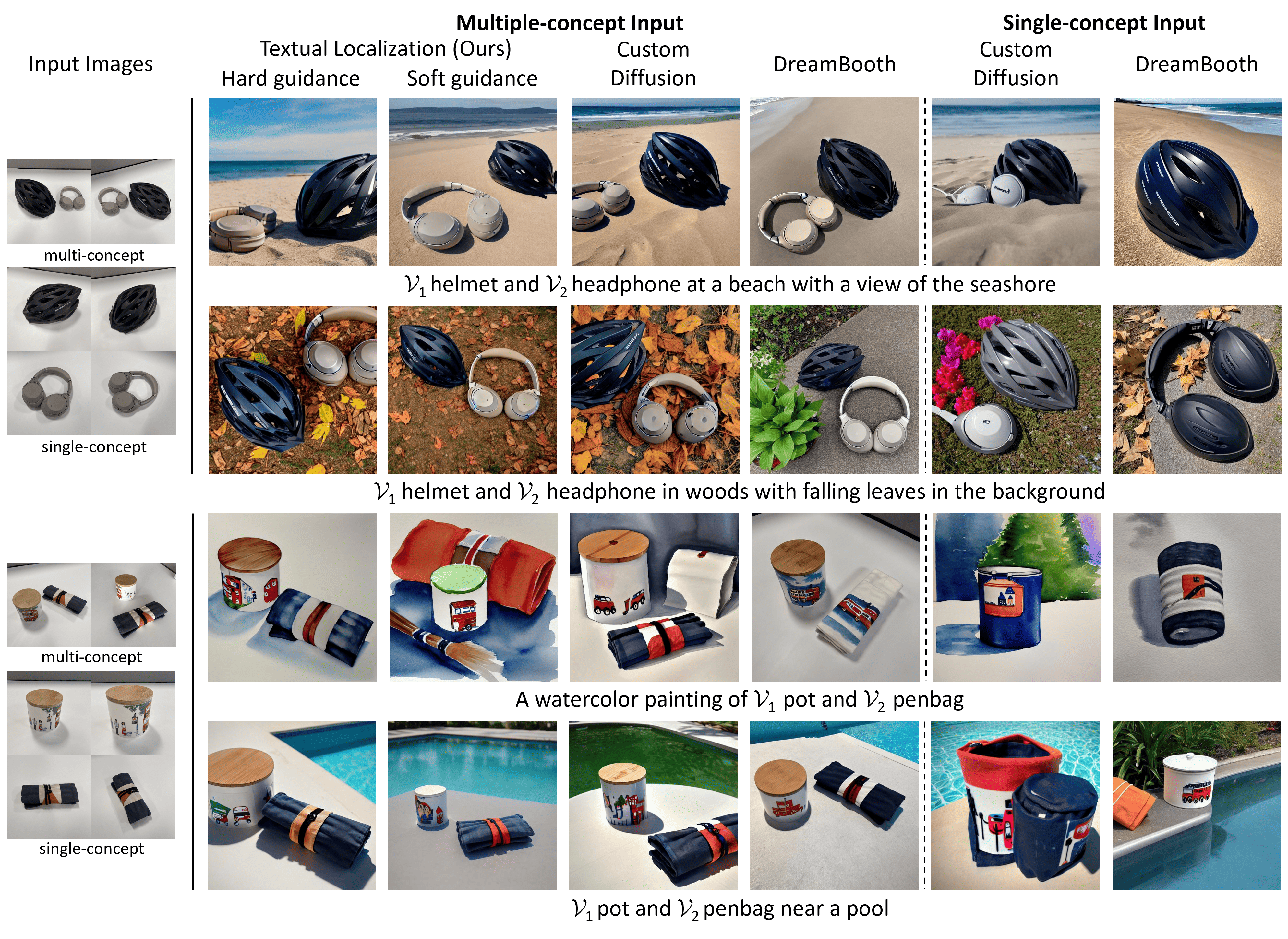}}
\caption{Qualitative comparison in multi-concept generation}
\label{multiple_subject_result}
\end{center}
\vskip -0.1in
\end{figure*}

\subsection{Experimental setup}
\label{sec4.1}

\textbf{Datasets.} We curated a dataset comprising 10 novel concepts encompassing general and everyday items. To assess the model's efficacy with multi-concept images, we randomly formed five groups by pairing two concepts. Additionally, we prepared single-concept images for each concept to facilitate evaluation. Each single and multiple concept set consists of five images, and samples can be found in \cref{appendixA}. To pinpoint the locations of the target concepts in the input images, we employed the Grounding DINO detection model \citep{liu2023grounding} to generate bounding boxes. These bounding boxes served as cues for the segmentation model SAM \citep{kirillov2023segment} to derive segmentation maps for the target concepts.

\textbf{Baseline models.} We conduct a comparative analysis of our method against two baseline models, namely, \textit{DreamBooth} \citep{ruiz2023dreambooth} and \textit{Custom Diffusion} \citep{kumari2023multi}. Both baseline models are subject-driven text-to-image models and integrate the class-specific prior preservation loss to mitigate language drift. \textit{DreamBooth} undertakes fine-tuning of the text encoder and the entire UNet, while \textit{Custom Diffusion} focuses solely on optimizing the text encoder and the $W_K$ and $W_V$ matrices within the cross-attention layers of the UNet.

\textbf{Evaluation metrics.} We evaluate our method on the benchmark raised by the baseline models to present a fair comparison, focusing on image fidelity and image-text alignment. To gauge image fidelity, we compute the cosine similarity between the CLIP embeddings of the generated images and the real images, denoted as CLIP-I. Additionally, we calculate the Kernel Inception Distance (KID) \citep{binkowski2018demystifying} between the generated and real images, providing further evidence of image fidelity. Furthermore, the average Learned Perceptual Image Patch Similarity (LPIPS) \citep{zhang2018unreasonable} is computed for the generated images of the same target concept under identical text prompts. This measure reveals the diversity of the generated images and aids in assessing potential model overfitting. For image-text alignment, the cosine similarity between CLIP embeddings of the generated images and the corresponding text prompts is calculated (CLIP-T). Note that when evaluating CLIP-T, the identifier token $\mathcal{V}$ is omitted, as the CLIP text encoder has not undergone fine-tuning on $\mathcal{V}$.

\textbf{Implementation details.} For all experiments, we employ SD v1.5 trained on the LAION-5B dataset \citep{schuhmann2022laion} as the foundational model. Prior to fine-tuning, we leverage the pre-trained SD to generate 200 images per target concept for computing $L_{prior}$, using text prompts consisting of the class name of the respective target concept. In the case of our method, the scaling coefficients $\lambda$ and $\delta$ are set to 1.0. We extract cross-attention maps from cross-attention layers with downsampling factors $f \in {2,4,8}$ \citep{hertz2022prompt, chefer2023attend}, and up-scale to $H' \times W'$, where $H' = W' = 256$ for computational efficiency. The identifier token $\mathcal{V}$ ($\mathcal{V}_1$ and $\mathcal{V}_2$ for multiple concepts) is initialized with the token ID 48136 in the pre-trained CLIP tokenizer \citep{gal2022image, ruiz2023dreambooth}. The learning rate is set to $1.0 \times 10^{-5}$ for our method and \textit{Custom Diffusion}, and $5.0 \times 10^{-6}$ for \textit{DreamBooth}. Maintaining a fixed batch size of 2, all models are fine-tuned on an NVIDIA A100 GPU.

\subsection{Single-concept generation}
\label{sec4.2}

We assess each model's capability to generate a single new concept when provided with multi-concept images as input. For each target concept, we employ 10 text prompts and generate 50 image samples per prompt, resulting in a total of 500 images. Additionally, we evaluate the performance of baseline models when taking single-concept images as input for a more comprehensive analysis. The average model performance across all target concepts is summarized in \cref{comparison}. Our method achieves superior performance in CLIP-T and LPIPS, with the soft guidance variant obtaining the highest score and the hard guidance variant securing the second-best score. This indicates that our method effectively preserves more semantic information from the text prompts and exhibits reduced overfitting. On the other hand, \textit{DreamBooth}, fine-tuned on single-concept images, attains the highest scores in CLIP-I and KID, reflecting higher similarities between input and generated images. However, it is noteworthy that these results may be biased, as baseline models fine-tuned on single-concept images also use the same images for evaluation. Moreover, \textit{DreamBooth}, optimizing more parameters, showcases enhanced learning of visual representation but also accelerates the loss of learned prior knowledge, resulting in lower CLIP-T and LPIPS scores. Furthermore, compared to \textit{Custom Diffusion}, our method exhibits an overall improvement, as the CLIP-I, CLIP-T, and LPIPS scores of the hard guidance variant are 7.04\%, 2.22\%, 0.91\% higher, and the KID score is 6.10\% lower on multi-concept images.

The qualitative evaluation results are presented in \cref{single_subject_result}. Our method, fine-tuned on multi-concept images, successfully generates images containing only the target concept while retaining rich semantic information from the text prompt. In contrast, both baseline models encounter difficulties in clarifying the target concept and are prone to generating images containing all concepts present in the input images. Additionally, while \textit{DreamBooth} demonstrates a better visual representation of the input concept, it tends to lose more semantic knowledge, leading to results inconsistent with the text prompts (e.g., $\mathcal{V}$ \textit{doll under water}). Further examples and analysis can be found in \cref{appendixB-1}.

\subsection{Multi-concept generation}
\label{sec4.3}

\begin{table*}[htb!]
\caption{Comparison between our method and baseline models (Bold indicates the best value, underline represents the second-best value)}
\label{comparison}
\vskip 0.1in
\begin{center}
\begin{small}
\begin{tabular}{lccccccccc}
\toprule
\multirow{2}{*}{Method} & \multirow{2}{*}{Input Concept} & \multicolumn{4}{c}{Single-concept generation} & \multicolumn{4}{c}{Multi-concept generation} \\ 
 & & CLIP-I $\uparrow$ & CLIP-T $\uparrow$ & KID $\downarrow$ & LPIPS $\uparrow$ & CLIP-I $\uparrow$ & CLIP-T $\uparrow$ & KID $\downarrow$ & LPIPS $\uparrow$ \\
\midrule
DreamBooth                &Single     & \textbf{0.6583}    & 0.2161             & \textbf{0.1297}    & 0.5884                        & \underline{0.5189} & 0.1970             & 0.1952             & 0.5971 \\
Custom Diffusion          &Single     & 0.5525             & 0.2645             & \underline{0.1865} & 0.6355                        & 0.4600             & \textbf{0.2890}    & 0.2179             & \underline{0.6485} \\
DreamBooth                &Multiple   & \underline{0.5999} & 0.2099             & 0.1985             & 0.6162                        & 0.5034             & 0.2136             & \underline{0.1625} & 0.6036 \\
Custom Diffusion          &Multiple   & 0.4883             & 0.2612             & 0.2228             & 0.6595                        & 0.4907             & 0.2548             & 0.1743             & 0.5890 \\
Ours (hard guidance)      &Multiple   & 0.5227             & \underline{0.2670} & 0.2092             & \underline{0.6655}            & \textbf{0.5306}    & \underline{0.2697} & \textbf{0.1608}    & 0.6332 \\
Ours (soft guidance)      &Multiple   & 0.5077             & \textbf{0.2680}    & 0.2205             & \textbf{0.6685}               & 0.4951             & 0.2638             & 0.1781             & \textbf{0.6508}  \\
\bottomrule
\end{tabular}
\end{small}
\end{center}
\vskip -0.1in
\end{table*}

We evaluate each model's performance in generating multiple new concepts when fine-tuning them on multi-concept images. Two identifier tokens, $\mathcal{V}_1$ and $\mathcal{V}_2$, are introduced in the text prompts to represent the two target concepts. All target concepts are learned simultaneously during fine-tuning, pairing the input images with the text prompt ``\textit{photo of a} $\mathcal{V}_1$ [$class_1$] \textit{and a} $\mathcal{V}_2$ [$class_2$]". Baseline models are also jointly fine-tuned on single-concept images of the two target concepts, following the approach by \citet{kumari2023multi}. We generate 50 images per prompt for 10 text prompts in each multi-concept group, resulting in a total of 500 images. Quantitative evaluation results are presented in \cref{comparison}. Overall, our method exhibits superior performance in multi-concept generation. The hard guidance variant achieves the best CLIP-I and KID scores as well we the second-best CLIP-T score, while the soft guidance variant records the highest LPIPS score. In comparison to \textit{Custom Diffusion}, the CLIP-I, CLIP-T, and LPIPS scores of the hard guidance variant are 8.13\%, 5.85\%, 7.50\% higher, and the KID score is 7.75\% lower on multi-concept images. It's important to note that as both fine-tuning and evaluation use multi-concept images, there might be bias in the results when comparing our method with baseline models fine-tuned on single-concept images. Nevertheless, our method outperforms both baseline models fine-tuned on multi-concept images in terms of CLIP-I, KID, and LPIPS. Notably, while \textit{Custom Diffusion} fine-tuned on single-concept images attains the highest CLIP-T score, it exhibits the worst CLIP-I and KID scores, indicating insufficient learning of visual representation from input images, resulting in less loss of semantic knowledge during fine-tuning.

Results of the qualitative evaluation in multi-concept generation are presented in \cref{multiple_subject_result}. All models demonstrate the capability to generate both target concepts when fine-tuned on multi-concept images. However, \textit{DreamBooth} exhibits a more pronounced loss of semantic knowledge, leading to weaker representations of text prompts in some sample images. For instance, with the text prompt ``$\mathcal{V}_1$ \textit{helmet and} $\mathcal{V}_2$ \textit{headphone at a beach with a view of the seashore}", only a small amount of water is presented. Similarly, a watercolor painting is shown on the penbag with the text prompt ``\textit{A watercolor painting of} $\mathcal{V}_1$ pot and $\mathcal{V}_2$ \textit{penbag}". Furthermore, baseline models trained on single-concept images display property fusion of the two target concepts or only depict one concept in generated images. In contrast, our method appropriately presents both target concepts while accurately reflecting semantic information from the text prompts. Additional examples and detailed results are provided in \cref{appendixB-2}.

\subsection{Probing into cross-attention maps}
\label{sec4.4}

\begin{figure}[htb!]
\vskip 0.1in
\begin{center}
\centerline{\includegraphics[width=1.0\columnwidth]{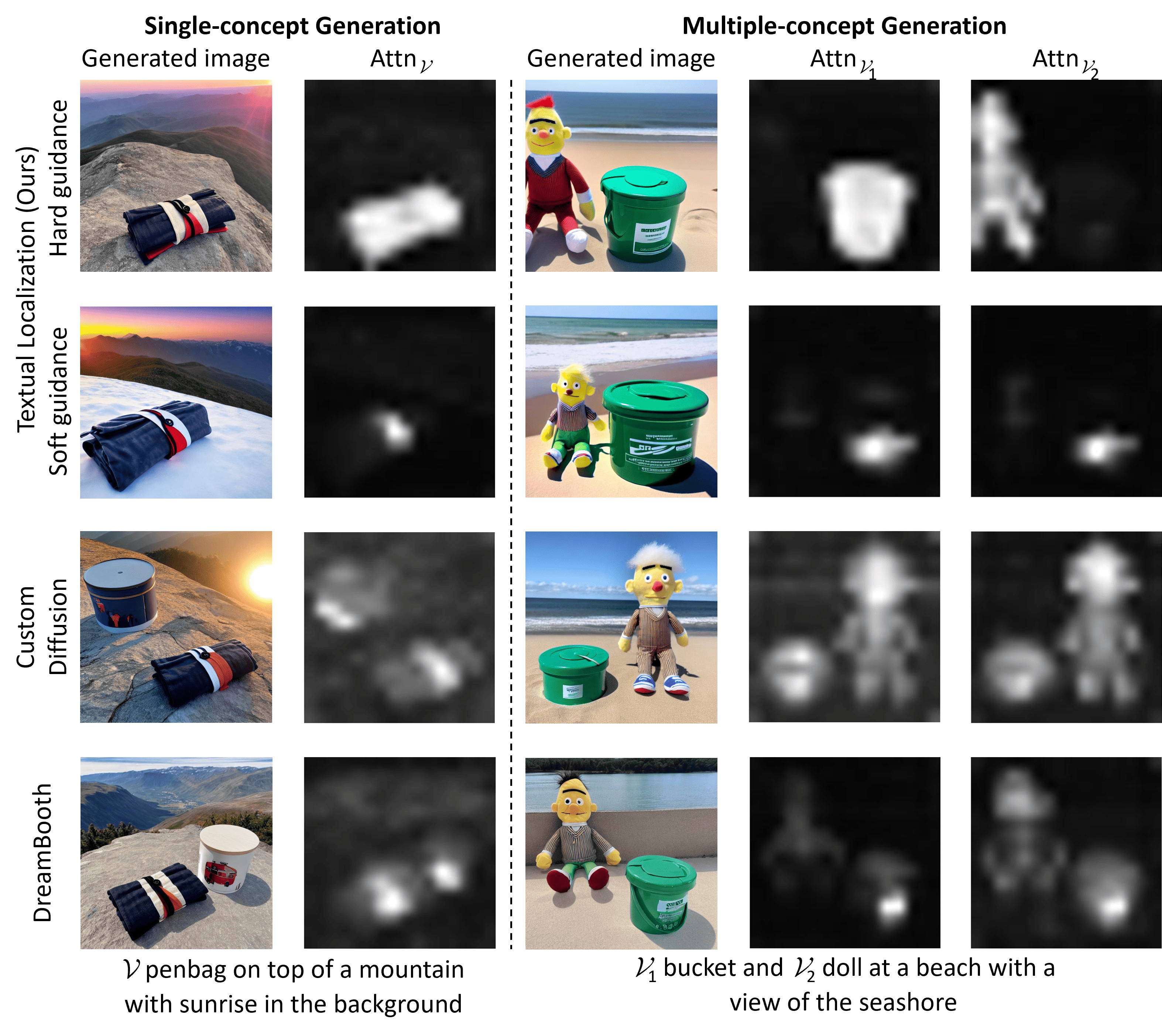}}
\caption{Images samples and cross-attention maps of identifier tokens generated by adopted models fine-tuned on multi-concept input images}
\label{cross-attention-map}
\end{center}
\vskip -0.1in
\end{figure}

A crucial aspect of subject-driven models is the establishment of a connection between the identifier token $\mathcal{V}$ and the visual representation of the target concept. However, this connection heavily relies on the diffusion models' perceptual ability to accurately locate the target region. When confronted with multi-concept images, the connection becomes ambiguous due to a lack of guidance, resulting in the presence of non-target concepts in the generated images. To showcase the connection, we extract cross-attention maps from the $16 \times 16$ ($f=4$) layers of the UNet throughout all timesteps during inference, and then upscale to $256 \times 256$, which are displayed in \cref{cross-attention-map}.

It can be observed that both \textit{DreamBooth} and \textit{Custom Diffusion} tend to produce the non-target concept in single-concept generation, and the cross-attention maps of the identifier token outline the shape of all concepts present in the generated images in both single and multiple concept generation. Moreover, a comparative analysis between the soft guidance variant and the hard guidance variant of our method shows that while both variants avoid generating the non-target concept, the former only partially depicts the outline of the target concept in the cross-attention maps in single-concept generation. This partial connection explains the higher CLIP-T score but lower CLIP-I score of the soft guidance variant compared to the hard guidance variant. In multi-concept generation, the soft guidance variant struggles to differentiate cross-attention maps for different concepts, whereas the hard guidance variant accurately depicts the outlines of each target concept. Consequently, ambiguities persist when using the soft guidance variant for multi-concept generation, leading to lower CLIP-I and CLIP-T scores compared to the hard guidance variant.

\subsection{Ablation study}
\label{sec4.5}
As detailed in \cref{comparison}, \textit{DreamBooth} achieves a higher CLIP-I score but a lower CLIP-T score compared to \textit{Custom Diffusion} when provided with the same input images as \textit{DreamBooth} optimizes more model parameters, resulting in a more pronounced loss of prior semantic knowledge. Determining an optimal set of trainable parameters is crucial for improving the model's ability to learn visual representation while retaining semantic knowledge. To explore this, we conducted an ablation study on parameter optimization.

\begin{figure}[htb!]
\vskip 0.1in
\begin{center}
\centerline{\includegraphics[width=1.0\columnwidth]{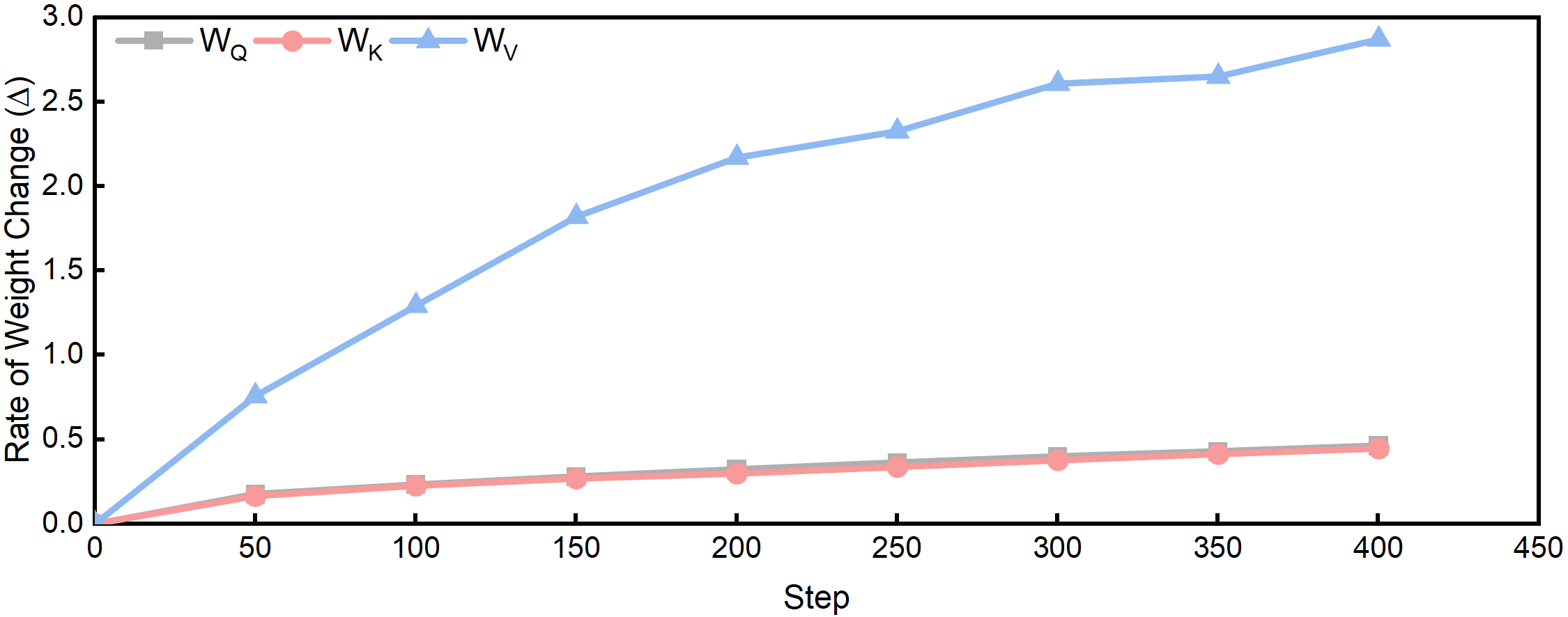}}
\caption{Rates of weight change of different matrices in cross-attention layers}
\label{weight-change}
\end{center}
\vskip -0.1in
\end{figure}

Building upon the conclusion by \citet{kumari2023multi}, we delve deeper by evaluating the rate of weight change for the $W_Q$, $W_K$, and $W_V$ matrices in the cross-attention layers. We fine-tuned the cross-attention layers on six single concepts from our dataset, calculating the mean rate of weight change for each layer using $\Delta_l = || \theta_{l}^{'} - \theta_{l} || / || \theta_{l} ||$, where $\theta_{l}$ and $\theta_{l}^{'}$ represent weights of the parameters in layer $l$ before and after fine-tuning \cite{li2020few}. The rates of weight change for different matrices with fine-tuning steps are presented in \cref{weight-change}. Notably, the weights of $W_V$ undergo the most significant change, while the weights of $W_Q$ and $W_K$ exhibit less pronounced changes. Given this observation, we select three sets of model parameters for optimization: (1) $W_Q + W_K + W_V$, (2) $W_Q + W_V$, and (3) $W_K + W_V$ (adopted in our method). We fine-tune these three parameter sets within the framework of our method with hard guidance and evaluate the generated images. \cref{ablation} showcases the results of the ablation study, revealing that optimizing $W_K + W_V$ achieves the best or second-best performance across most metrics in single-concept generation. Remarkably, it attains both the highest CLIP-I and CLIP-T scores in multi-concept generation. Therefore, optimizing $W_K + W_V$ emerges as a rational choice in this study. Additional details are provided in \cref{appendixB-3}.

\begin{table}[htb!]
\caption{Ablation study of different selections of trainable parameters (Bold indicates the best value, underline represents the second-best value)}
\label{ablation}
\vskip 0.1in
\begin{center}
\begin{small}
\begin{tabularx}{1.0\columnwidth}{ >{\raggedright\arraybackslash}p{2.15cm}cccc}
\toprule
Parameter Set & CLIP-I $\uparrow$ & CLIP-T $\uparrow$ & KID $\downarrow$ & LPIPS $\uparrow$ \\
\midrule
\multicolumn{5}{l}{\textbf{Single-concept generation}} \\
$W_Q + W_K + W_V$ & \textbf{0.5288}    & 0.2629             & \textbf{0.1985}    & 0.6587 \\
$W_Q + W_V$       & 0.5142             & \underline{0.2658} & 0.2103             & \underline{0.6601} \\
$W_K + W_V$       & \underline{0.5227} & \textbf{0.2670}    & \underline{0.2092}    & \textbf{0.6655} \\
\midrule
\multicolumn{5}{l}{\textbf{Multi-concept generation}} \\
$W_Q + W_K + W_V$   & \underline{0.5224}  & 0.2644 & \textbf{0.1512}     & \underline{0.6313} \\
$W_Q + W_V$         & 0.5133              & \underline{0.2659}             & \underline{0.1583}  & 0.6303 \\
$W_K + W_V$         & \textbf{0.5306}     & \textbf{0.2697}    & 0.1608              & \textbf{0.6332} \\
\bottomrule
\end{tabularx}
\end{small}
\end{center}
\vskip -0.1in  
\end{table}

\section{Discussion and Conclusion}

\begin{figure}[htb!]
\vskip 0.1in
\begin{center}
\centerline{\includegraphics[width=1.0\columnwidth]{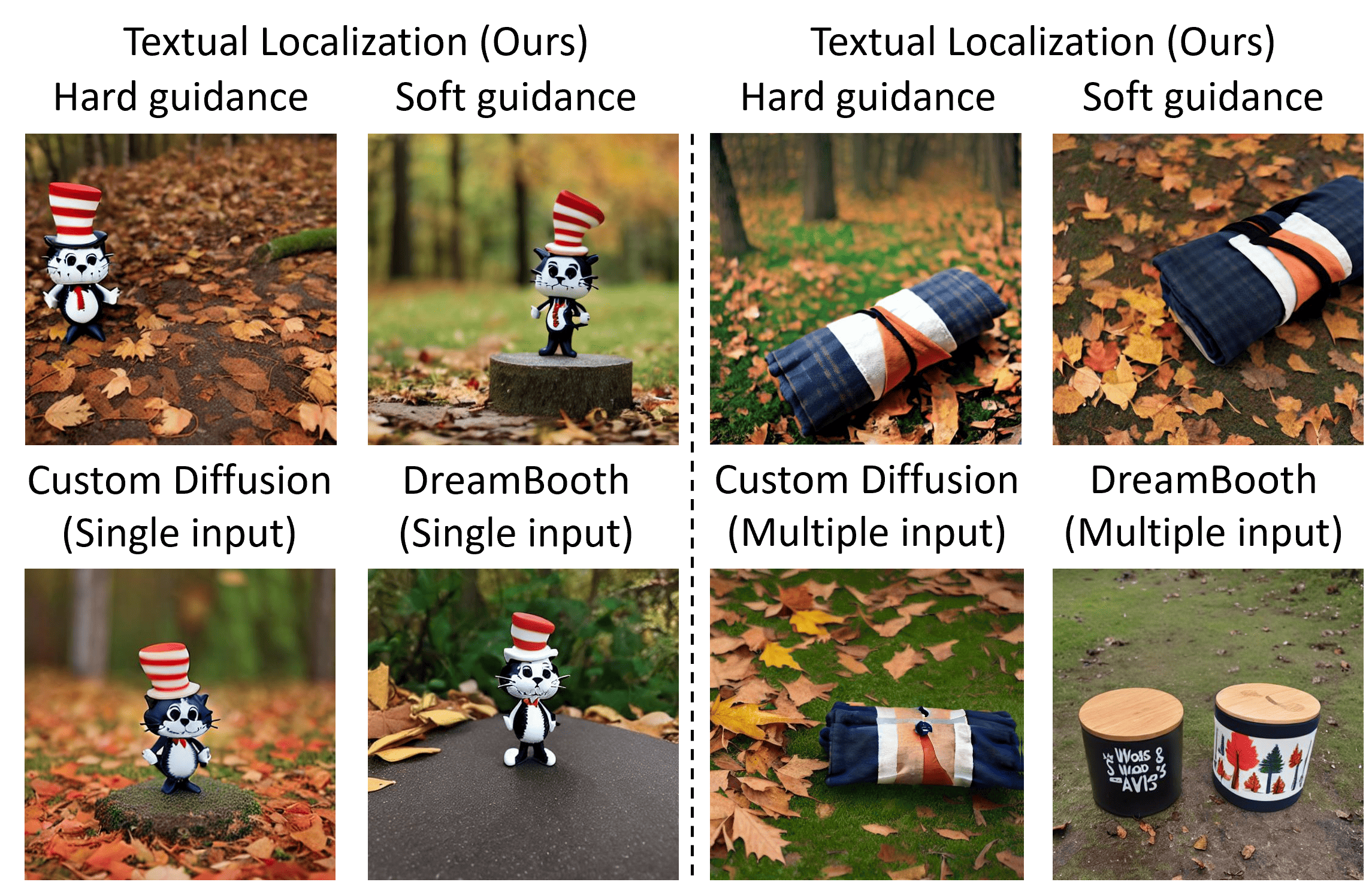}}
\caption{Failure cases on subject-driven text-to-image generation. Left: failure to capture the details of the cat figurine; Right: failure to generate all concepts in the text prompt. (Text prompts: $\mathcal{V}_1$ \textit{cat figurine} (left) / $\mathcal{V}_1$ \textit{pot and} $\mathcal{V}_2$ \textit{penbag} (right) \textit{in woods with falling leaves in the background})}
\label{fail-case}
\end{center}
\vskip -0.1in
\end{figure}

This study introduces a novel subject-driven text-to-image model, termed \textit{Textual Localization}, aimed at mitigating ambiguities inherent in subject-driven models on multi-concept input images. The proposed method incorporates a novel cross-attention guidance to disentangle multiple concepts from input images and establish accurate connections between the visual representations of the target concept and the identifier token in the text prompt. Our method demonstrates superior or comparable performance to baseline models in terms of image fidelity and image-text alignment on multi-concept input images, as the hard guidance variant achieves CLIP-I scores that are 7.04\%, 8.13\% higher, and CLIP-T scores that are 2.22\%, 5.85\% higher than \textit{Custom Diffusion} in single-concept and multi-concept generation, respectively. Notably, our technique effectively delineates the outlines of target concepts in cross-attention maps.

Nevertheless, limitations emerge in capturing intricate details of target concepts as shown in \cref{fail-case}. Additionally, failure cases may arise in multi-concept generation, where only one concept is generated despite models being fine-tuned on multi-concept images, as depicted in \cref{fail-case}. Hence, our focus in future work will be on addressing these limitations. Specifically, we propose adopting more powerful feature extractors \cite{chen2023anydoor} to accentuate details in the input images, and integrating guiding techniques during inference \cite{chefer2023attend, chen2023training} to ensure the successful generation of all target concepts mentioned in the text prompt.

\newpage



\section*{Impact Statements}
This paper presents work whose goal is to advance the field of Machine Learning. There are many potential societal consequences of our work, none of which we feel must be specifically highlighted here.


\bibliography{main}
\bibliographystyle{icml2024}

\newpage
\appendix
\onecolumn

{\Large \textbf{Appendix}}

\section{Self-constructed Dataset}
\label{appendixA}

\begin{figure*}[htb!]
\vskip 0.1in
\begin{center}
\centerline{\includegraphics[width=0.8\columnwidth]{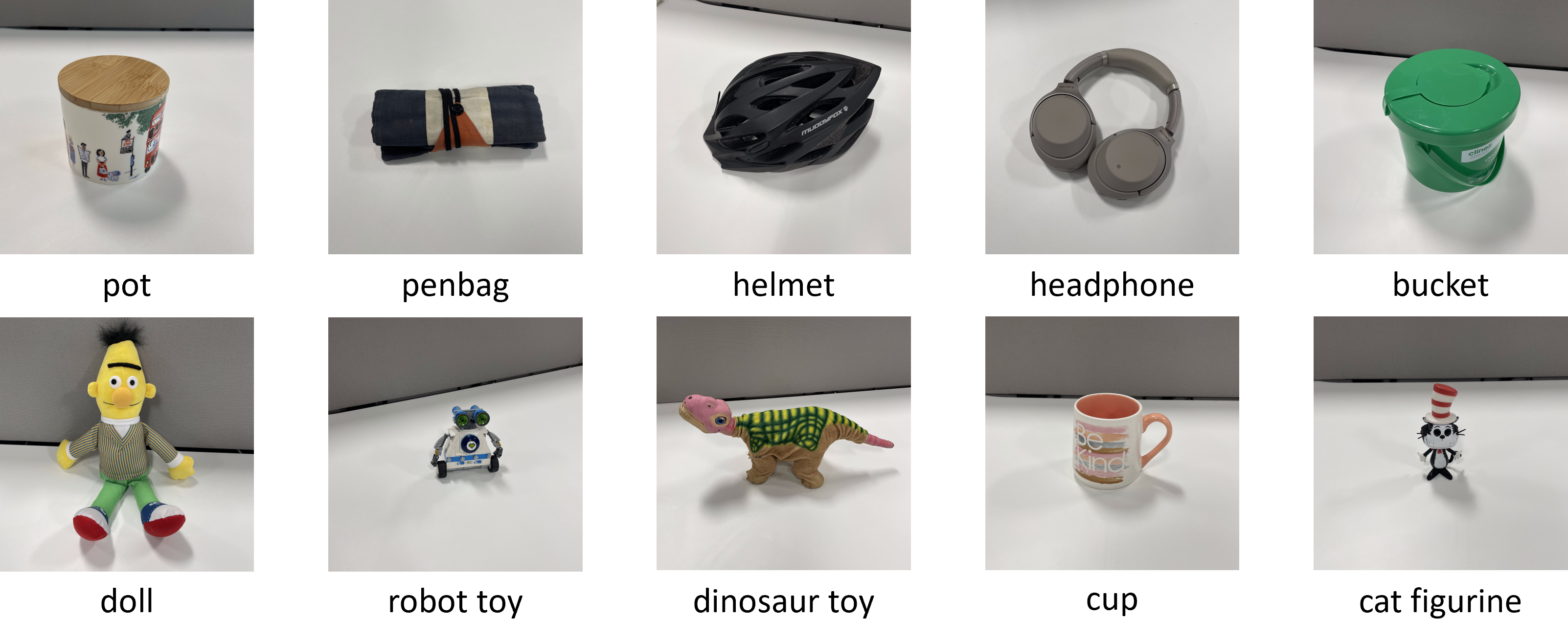}}
\caption{Samples of the single-concept images of each concept in the dataset}
\label{dataset_single}
\end{center}
\vskip -0.1in
\end{figure*}

\begin{figure*}[htb!]
\vskip 0.1in
\begin{center}
\centerline{\includegraphics[width=0.8\columnwidth]{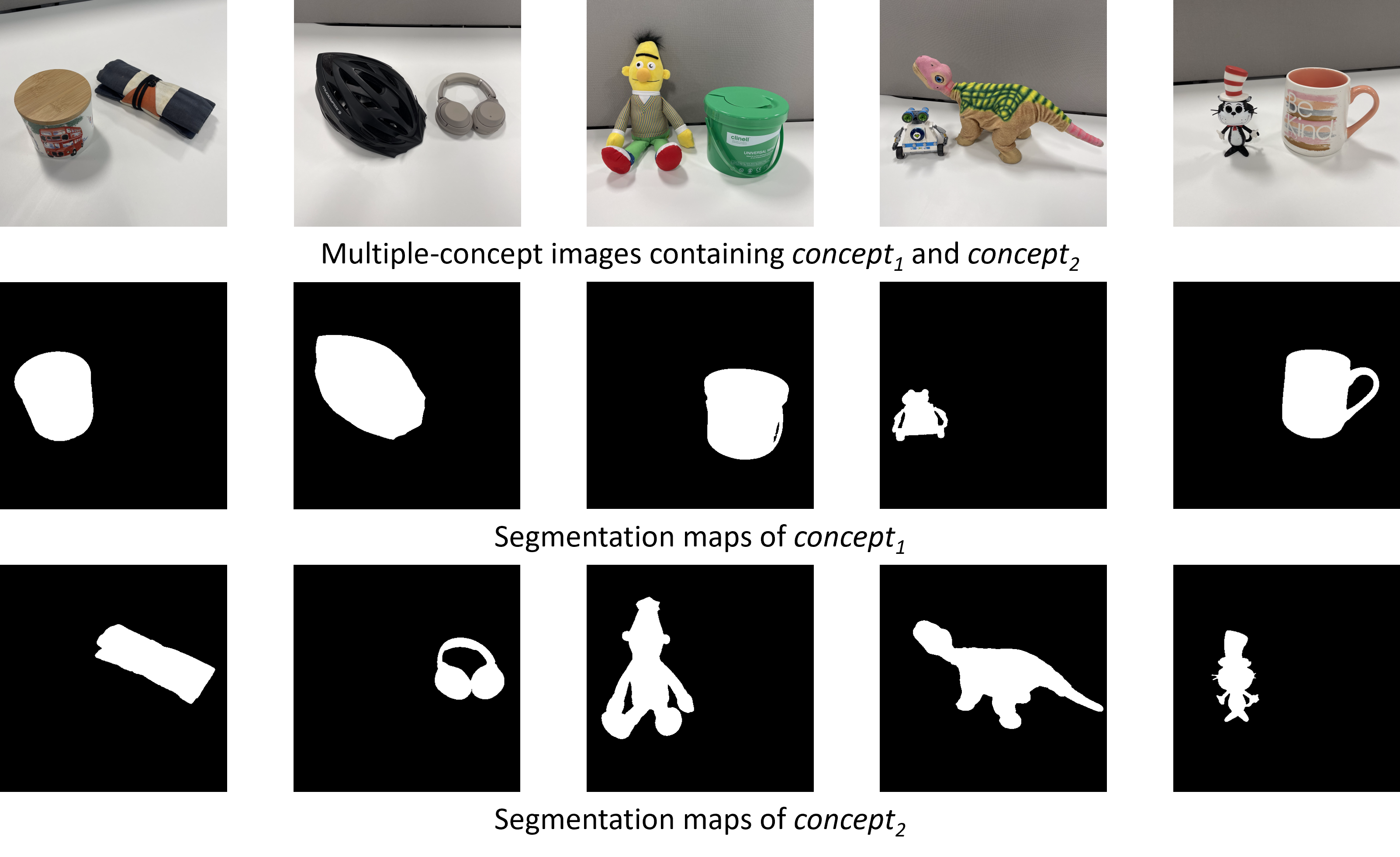}}
\caption{Samples of the multi-concept images with segmentation maps of individual concepts}
\label{dataset_multi}
\end{center}
\vskip -0.1in
\end{figure*}

Our experiments encompass evaluations on both single-concept and multi-concept images. Existing datasets for subject-driven text-to-image models, such as the DreamBooth Dataset \cite{ruiz2023dreambooth} and CustomConcept101 \cite{kumari2023multi}, primarily comprise single-concept images. However, the availability of datasets featuring multi-concept images is limited. To address this, we curated a self-constructed dataset containing both single-concept and multi-concept images, each depicting various general and everyday items. The dataset comprises 10 concepts, each accompanied by 5 single-concept images, as illustrated in \cref{dataset_single} along with their corresponding class names. For multi-concept images, we randomly grouped the concepts into 5 pairs, collecting 5 images for each pair. Additionally, we employed a pre-trained detection model, Grounding DINO \cite{liu2023grounding}, to identify the locations of each concept in the input images, generating corresponding bounding boxes. These bounding boxes served as input for a pre-trained segmentation model, SAM \cite{kirillov2023segment}, which produced segmentation maps for each concept. Samples of the multi-concept images and the segmentation maps for individual concepts are displayed in \cref{dataset_multi}.

\section{Complementary Experimental Results}
\label{appendixB}

Due to space constraints, only a summary of the experimental results is provided in \cref{experiments}. For a more in-depth examination, detailed experimental results and analysis are presented in this section.

\subsection{Single-concept generation}
\label{appendixB-1}

\begin{figure*}[htb!]
\vskip 0.1in
\begin{center}
\centerline{\includegraphics[width=1.0\columnwidth]{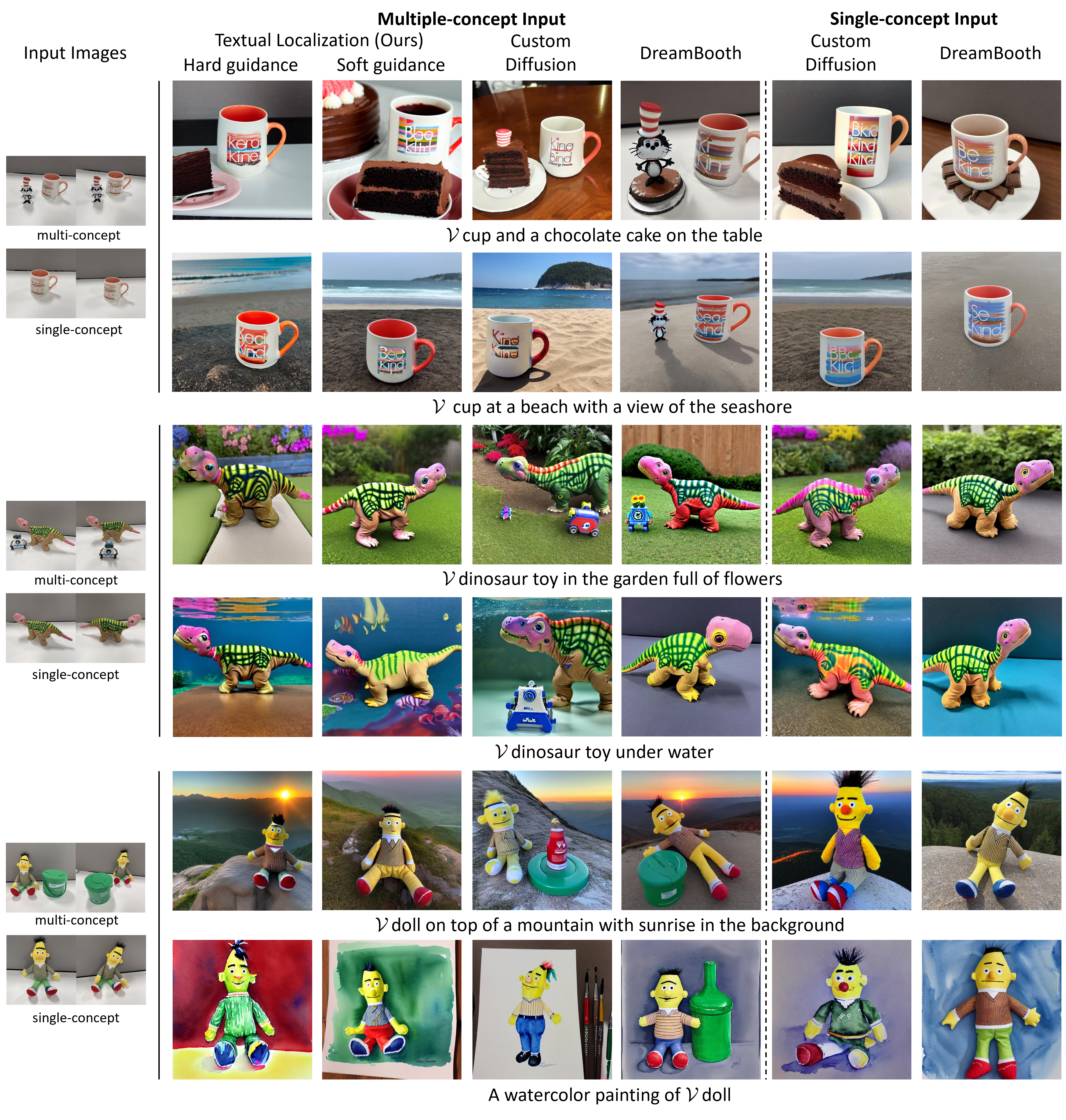}}
\caption{Complementary result of qualitative comparison in single-concept generation}
\label{appendix-single-concept}
\end{center}
\vskip -0.1in
\end{figure*}

The CLIP-I, CLIP-T, KID, and LPIPS scores for our method and the baseline models on each single concept are detailed in \cref{appendix-comparison-clip-i} to \cref{appendix-comparison-lpips}. Notably, \textit{DreamBooth} fine-tuned on single-concept images achieves the highest image fidelity, securing the top CLIP-I score on all single concepts except for the headphone (where it obtains the second-best score) and the lowest KID score across all single concepts. Interestingly, \textit{DreamBooth} fine-tuned on multi-concept images obtains the second-best CLIP-I and KID scores. This suggests that fine-tuning more parameters can be advantageous for learning visual representation from input images. However, a trade-off between acquiring visual representation and losing prior semantic knowledge is evident, as both variants of \textit{DreamBooth} generally yield lower CLIP-T and LPIPS scores compared to other methods. As evident in \cref{single_subject_result} and \cref{appendix-single-concept}, \textit{DreamBooth} struggles to fully capture semantic information from text prompts.

Comparatively, when assessing our method against \textit{Custom Diffusion}, both the hard guidance variant and soft guidance variant outperform \textit{Custom Diffusion} fine-tuned on multi-concept images across all evaluation metrics. This superiority can be attributed to \textit{Custom Diffusion} having a tendency to generate all concepts from multi-concept images, impacting the image fidelity and image-text alignment of the generated images. Examples of generating all input concepts are illustrated in \cref{single_subject_result} and \cref{appendix-single-concept}. In contrast, our method can precisely specify the target concept, resulting in improved performance. While the CLIP-I and KID scores of our method closely align with those of \textit{Custom Diffusion} fine-tuned on single-concept images, the CLIP-T and LPIPS scores are higher.

\subsection{Multi-concept generation}
\label{appendixB-2}

The CLIP-I, CLIP-T, KID, and LPIPS scores for our method and the baseline models on multi-concept groups are detailed in \cref{appendix-comparison-clip-i} to \cref{appendix-comparison-lpips}. In multi-concept generation, the hard guidance variant of our method achieves superior image fidelity compared to the baseline models in three groups. Notably, the hard guidance variant outperforms the soft guidance variant on all multi-concept groups, suggesting that hard guidance can introduce more visual representation to the model. The cross-attention map in \cref{cross-attention-map} further illustrates that only the hard guidance variant aligns with the position of the target concepts in generated images, whereas the soft guidance variant outlines only a part of the target concept, and the attention distribution does not correspond to the respective concepts in multi-concept generation. Hence, we conclude that hard guidance is favorable for multi-concept generation.

\begin{figure*}[htb!]
\vskip 0.1in
\begin{center}
\centerline{\includegraphics[width=1.0\columnwidth]{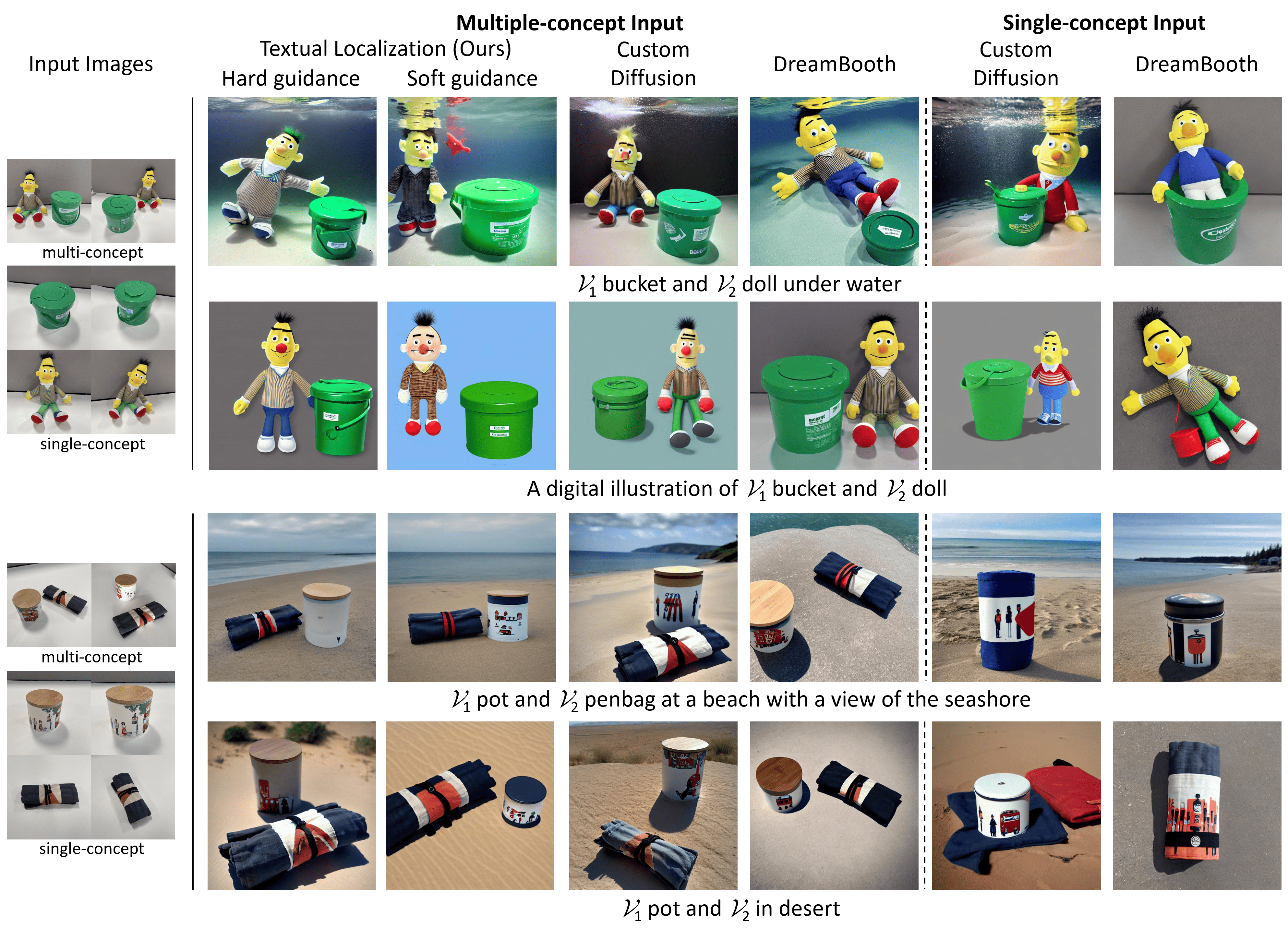}}
\caption{Complementary result of qualitative comparison in multi-concept generation}
\label{appendix-multi-concept}
\end{center}
\vskip -0.1in
\end{figure*}

Additionally, \textit{Custom Diffusion} trained on single-concept images achieves the highest CLIP-T score, but its image fidelity is low. As evident in \cref{multiple_subject_result} and \cref{appendix-multi-concept}, properties of the two input concepts are infused in the generated images. This property infusion is also observed in the images generated by \textit{DreamBooth} fine-tuned on single-concept images. It's worth noting that property infusion does not always occur when jointly training on single-concept images, as evidenced by some successful cases presented in \cref{appendix-multi-concept}. However, for better results in multi-concept generation, we recommend training on multi-concept images using our method, which avoids property infusion and enhances generation quality.

\begin{table*}[htb!]
\caption{Comparison of CLIP-I value between our method and baseline models on each concept in the dataset (Bold indicates the best value, underline represents the second-best value)}
\label{appendix-comparison-clip-i}
\vskip 0.1in
\begin{center}
\begin{small}
\begin{tabular}{lcccccc}
\toprule
\multirow{3}{*}{Target Concept} & \multicolumn{4}{c}{Multi-concept Input} & \multicolumn{2}{c}{Single-concept Input} \\
                                & Ours  & Ours & Custom & \multirow{2}{*}{DreamBooth} &  Custom & \multirow{2}{*}{DreamBooth} \\
                                & (hard guidance) & (soft guidance)  & Diffusion &    &  Diffusion & \\
\midrule
pot          & 0.4191 & 0.4114 & 0.4355 & \underline{0.6192} & 0.5002 & \textbf{0.7431} \\
penbag       & 0.4839 & 0.4852 & 0.4673 & \underline{0.6123} & 0.4950 & \textbf{0.6522} \\
helmet       & 0.5619 & 0.5514 & 0.5711 & \underline{0.6598} & 0.6037 & \textbf{0.6743} \\
headphone    & 0.6783 & 0.6347 & 0.5229 & \textbf{0.7290} & 0.6954 & \underline{0.6901} \\
bucket       & 0.4420 & 0.4529 & 0.3990 & \underline{0.5287} & 0.4618 & \textbf{0.5657} \\
doll         & 0.5012 & 0.4575 & 0.4421 & \underline{0.6091} & 0.5456 & \textbf{0.6979} \\
robot toy    & 0.5167 & 0.5285 & 0.5073 & \underline{0.5148} & 0.5558 & \textbf{0.5905} \\
dinosaur toy & 0.6433 & 0.6257 & 0.6010 & \underline{0.7060} & 0.6470 & \textbf{0.7270} \\
cup          & 0.4210 & 0.4059 & 0.4187 & \underline{0.4677} & 0.4601 & \textbf{0.6270} \\
cat figurine & 0.5592 & 0.5234 & 0.5181 & \underline{0.5531} & 0.5608 & \textbf{0.6122} \\
\midrule
pot \& penbag             & 0.4548 & 0.4504 & 0.4538 & \textbf{0.5913} & 0.4182 & \underline{0.5189} \\
helmet \& headphone       & 0.4964 & 0.4493 & 0.4729 & \textbf{0.5500} & 0.4296 & \underline{0.5227} \\
bucket \& doll            & \textbf{0.6006} & 0.5557 & \underline{0.5791} & 0.5236 & 0.5072 & 0.5556 \\
robot toy \& dinosaur toy & \textbf{0.5583} & 0.5201 & 0.5174 & 0.3931 & \underline{0.5222} & 0.5150 \\
cup \& cat figurine       & \textbf{0.5427} & \underline{0.4999} & 0.4304 & 0.4590 & 0.4285 & 0.4776 \\

\bottomrule
\end{tabular}
\end{small}
\end{center}
\vskip -0.1in
\end{table*}


\begin{table*}[htb!]
\caption{Comparison of CLIP-T value between our method and baseline models on each concept in the dataset (Bold indicates the best value, underline represents the second-best value)}
\label{appendix-comparison-clip-t}
\vskip 0.1in
\begin{center}
\begin{small}
\begin{tabular}{lcccccc}
\toprule
\multirow{3}{*}{Target Concept} & \multicolumn{4}{c}{Multi-concept Input} & \multicolumn{2}{c}{Single-concept Input} \\
                                & Ours  & Ours & Custom & \multirow{2}{*}{DreamBooth} &  Custom & \multirow{2}{*}{DreamBooth} \\
                                & (hard guidance) & (soft guidance)  & Diffusion &    &  Diffusion & \\
\midrule
pot          & \textbf{0.2354} & 0.2328 & 0.2267 & 0.1611 & \underline{0.2341} & 0.1498 \\
penbag       & 0.2478 & \textbf{0.2628} & 0.2354 & 0.2054 & \underline{0.2492} & 0.2191 \\
helmet       & \underline{0.2795} & \textbf{0.2800} & 0.2752 & 0.2175 & 0.2617 & 0.2214 \\
headphone    & \underline{0.2658} & \textbf{0.2675} & 0.2519 & 0.2266 & 0.2612 & 0.2319 \\
bucket       & 0.2798 & \textbf{0.2827} & 0.2781 & 0.2504 & \underline{0.2815} & 0.2583 \\
doll         & \underline{0.2413} & \textbf{0.2532} & 0.2398 & 0.1591 & 0.2346 & 0.1549 \\
robot toy    & \underline{0.2921} & 0.2884 & 0.2904 & 0.2230 & \textbf{0.2923} & 0.2615 \\
dinosaur toy & \underline{0.3008} & 0.2991 & 0.2957 & 0.2679 & \textbf{0.3066} & 0.2614 \\
cup          & \underline{0.2523} & 0.2475 & 0.2449 & 0.2129 & \textbf{0.2540} & 0.2034 \\
cat figurine & \textbf{0.2748} & 0.2658 & 0.2745 & 0.1754 & \underline{0.2700} & 0.1993 \\
\midrule
pot \& penbag             & \underline{0.2451} & 0.2369 & 0.2265 & 0.1913 & \textbf{0.2684} & 0.2106 \\
helmet \& headphone       & \underline{0.2765} & 0.2697 & 0.2713 & 0.2062 & \textbf{0.2918} & 0.1570 \\
bucket \& doll            & \underline{0.2621} & 0.2604 & 0.2551 & 0.1588 & \textbf{0.2671} & 0.1939 \\
robot toy \& dinosaur toy & \underline{0.2847} & 0.2796 & 0.2808 & 0.2300 & \textbf{0.3074} & 0.2269 \\
cup \& cat figurine       & 0.2803 & 0.2726 & 0.2407 & \underline{0.2819} & \textbf{0.3106} & 0.1970 \\

\bottomrule
\end{tabular}
\end{small}
\end{center}
\vskip -0.1in
\end{table*}


\begin{table*}[htb!]
\caption{Comparison of KID value between our method and baseline models on each concept in the dataset (Bold indicates the best value, underline represents the second-best value)}
\label{appendix-comparison-kid}
\vskip 0.1in
\begin{center}
\begin{small}
\begin{tabular}{lcccccc}
\toprule
\multirow{3}{*}{Target Concept} & \multicolumn{4}{c}{Multi-concept Input} & \multicolumn{2}{c}{Single-concept Input} \\
                                & Ours  & Ours & Custom & \multirow{2}{*}{DreamBooth} &  Custom & \multirow{2}{*}{DreamBooth} \\
                                & (hard guidance) & (soft guidance)  & Diffusion &    &  Diffusion & \\
\midrule
pot          & 0.1695 & 0.2026 & 0.1609 & \underline{0.0737} & 0.1621 & \textbf{0.0528} \\
penbag       & 0.2269 & 0.2326 & 0.2118 & \underline{0.1347} & 0.1984 & \textbf{0.0877} \\
helmet       & 0.1305 & 0.1505 & 0.1328 & 0.1473 & \underline{0.1163} & \textbf{0.0618} \\
headphone    & 0.2173 & 0.2462 & 0.2885 & \underline{0.1268} & 0.1306 & \textbf{0.1257} \\
bucket       & 0.2171 & 0.2165 & 0.2337 & \underline{0.1718} & 0.1991 & \textbf{0.1672} \\
doll         & 0.1515 & 0.1565 & 0.1601 & \underline{0.1010} & 0.1270 & \textbf{0.0724} \\
robot toy    & 0.3108 & 0.3083 & 0.3124 & 0.3501 & \underline{0.2739} & \textbf{0.2468} \\
dinosaur toy & 0.1554 & 0.1410 & 0.1709 & \underline{0.1200} & 0.1505 & \textbf{0.0964} \\
cup          & 0.0895 & 0.1105 & 0.1029 & 0.1152 & \underline{0.0954} & \textbf{0.0537} \\
cat figurine & 0.4239 & 0.4406 & 0.4544 & 0.6443 & \underline{0.4120} & \textbf{0.3320} \\
\midrule
pot \& penbag             & 0.2259 & 0.2541 & \underline{0.2170} & \textbf{0.1195} & 0.3051 & 0.2612 \\
helmet \& headphone       & 0.1174 & 0.1155 & \underline{0.1140} & \textbf{0.1098} & 0.1666 & 0.1132 \\
bucket \& doll            & 0.1088 & \underline{0.1028} & 0.1125 & \textbf{0.0890} & 0.1436 & 0.1370 \\
robot toy \& dinosaur toy & \textbf{0.2484} & 0.3074 & \underline{0.2942} & 0.3788 & 0.3302 & 0.3389 \\
cup \& cat figurine       & \textbf{0.1039} & \underline{0.1110} & 0.1339 & 0.1156 & 0.1438 & 0.1255 \\

\bottomrule
\end{tabular}
\end{small}
\end{center}
\vskip -0.1in
\end{table*}


\begin{table*}[htb!]
\caption{Comparison of LPIPS value between our method and baseline models on each concept in the dataset (Bold indicates the best value, underline represents the second-best value)}
\label{appendix-comparison-lpips}
\vskip 0.1in
\begin{center}
\begin{small}
\begin{tabular}{lcccccc}
\toprule
\multirow{3}{*}{Target Concept} & \multicolumn{4}{c}{Multi-concept Input} & \multicolumn{2}{c}{Single-concept Input} \\
                                & Ours  & Ours & Custom & \multirow{2}{*}{DreamBooth} &  Custom & \multirow{2}{*}{DreamBooth} \\
                                & (hard guidance) & (soft guidance)  & Diffusion &    &  Diffusion & \\
\midrule

pot          & 0.6578 & \textbf{0.6702} & \underline{0.6697} & 0.6059 & 0.6456 & 0.4382 \\
penbag       & \textbf{0.6669} & 0.6631 & \underline{0.6659} & 0.6598 & 0.6488 & 0.6229 \\
helmet       & \textbf{0.6651} & \underline{0.6544} & 0.6543 & 0.6021 & 0.6313 & 0.6505 \\
headphone    & 0.6172 & \underline{0.6266} & \textbf{0.6349} & 0.5825 & 0.5583 & 0.4841 \\
bucket       & \textbf{0.6968} & \underline{0.6909} & 0.6874 & 0.6894 & 0.6669 & 0.6407 \\
doll         & \underline{0.7006}& \textbf{0.7037} & 0.6896 & 0.6819 & 0.6776 & 0.6596 \\
robot toy    & \underline{0.6802} & \textbf{0.6833} & 0.6561 & 0.6085 & 0.6336 & 0.5734 \\
dinosaur toy & 0.6130 & \textbf{0.6338} & 0.6109 & \underline{0.6308} & 0.6278 & 0.6022 \\
cup          & \underline{0.6805} & \textbf{0.6841} & 0.6803 & 0.6186 & 0.6572 & 0.5958 \\
cat figurine & \textbf{0.6767} & \underline{0.6744} & 0.6461 & 0.4825 & 0.6084 & 0.6169 \\
\midrule
pot \& penbag             & \textbf{0.6672} & \underline{0.6643} & 0.6459 & 0.5514 & 0.6523 & 0.6613 \\
helmet \& headphone       & 0.6662 & \underline{0.6710} & 0.6636 & 0.5854 & \textbf{0.6780} & 0.6566 \\
bucket \& doll            & 0.5816 & 0.6204 & 0.5941 & \textbf{0.6301} & \underline{0.6268} & 0.5292 \\
robot toy \& dinosaur toy & 0.6242 & \underline{0.6448} & 0.6162 & \textbf{0.6464} & 0.6318 & 0.6258 \\
cup \& cat figurine       & 0.6268 & \textbf{0.6539} & 0.4252 & 0.6048 & \underline{0.6536} & 0.5129 \\

\bottomrule
\end{tabular}
\end{small}
\end{center}
\vskip -0.1in
\end{table*}


\subsection{Ablation study}
\label{appendixB-3}

In the ablation study, we explore the performance of our method when optimizing different parameter sets to identify the optimal trainable parameter set. Specifically, we test three parameter sets from the cross-attention layers in the UNet: $W_Q + W_K + W_V$, $W_Q + W_V$, and $W_K + W_V$. In the cross-attention layers, the $W_Q$ matrix handles information from the image, while the $W_K$ and $W_V$ matrices deal with information from text prompts. Moreover, the $W_Q$ and $W_K$ matrices are involved in the calculation of cross-attention maps, while the $W_V$ matrix carries the most semantic information from text prompts. Consequently, the rates of weight change of the $W_V$ matrix are most significant, as shown in \cref{weight-change}. The values of CLIP-I, CLIP-T, KID, and LPIPS metrics for each experiment are presented in \cref{appendix-ablation-clipi-clipt} and \cref{appendix-ablation-kid-lpips}.

The detailed results reveal that optimizing $W_Q + W_K + W_V$ is advantageous for achieving better image fidelity, as it obtains the best CLIP-I and KID scores for most target concepts. However, its performance on CLIP-T and LPIPS is inferior to the other sets. This observation aligns with the comparison between \textit{DreamBooth} and \textit{Custom Diffusion}, highlighting the trade-off between the acquisition of visual representations and the loss of prior semantic knowledge. Furthermore, there is no significant theoretical distinction between applying $W_Q + W_V$ or $W_K + W_V$. However, based on the experimental results, $W_K + W_V$ yields better performance than $W_Q + W_V$. Considering the overall performance across all evaluation metrics, $W_K + W_V$ is deemed the optimal choice for trainable parameters.

\begin{table}[htb!]
\caption{Comparison of CLIP-I and CLIP-T value between different parameter sets for optimization (Bold indicates the best value, underline represents the second-best value)}
\label{appendix-ablation-clipi-clipt}
\vskip 0.1in
\begin{center}
\begin{small}
\begin{tabular}{lcccccc}
\toprule
\multirow{2}{*}{Target Concept} & \multicolumn{3}{c}{CLIP-I $\uparrow$} & \multicolumn{3}{c}{CLIP-T $\uparrow$} \\
                                & $W_Q + W_K + W_V$ & $W_Q + W_V$ & $W_K + W_V$ & $W_Q + W_K + W_V$ & $W_Q + W_V$ & $W_K + W_V$ \\
\midrule

pot          & \textbf{0.4313} & \underline{0.4201} & 0.4191 & \underline{0.2345} & 0.2335 & \textbf{0.2354} \\
penbag       & \underline{0.4930} & \textbf{0.4934} & 0.4839 & 0.2405 & \textbf{0.2523} & \underline{0.2478} \\
helmet       & \underline{0.5573} & 0.5557 & \textbf{0.5619} & 0.2722 & \underline{0.2791} & \textbf{0.2795} \\
headphone    & \underline{0.6674} & 0.6483 & \textbf{0.6783} & 0.2626 & \underline{0.2629} & \textbf{0.2658} \\
bucket       & \underline{0.4400} & 0.4398 & \textbf{0.4420} & 0.2780 & \underline{0.2797} & \textbf{0.2798} \\
doll         & \textbf{0.5149} & 0.4989 & \underline{0.5012} & \textbf{0.2447} & \underline{0.2441} & 0.2413 \\
robot toy    & \underline{0.5247} & \textbf{0.5345} & 0.5167 & \textbf{0.2982} & \underline{0.2951} & 0.2921 \\
dinosaur toy & \textbf{0.6495} & 0.6410 & \underline{0.6433} & 0.2948 & \underline{0.2999} & \textbf{0.3008} \\
cup          & \textbf{0.4365} & 0.4010 & \underline{0.4210} & 0.2428 & \underline{0.2479} & \textbf{0.2523} \\
cat figurine & \textbf{0.5732} & 0.5091 & \underline{0.5592} & 0.2602 & \underline{0.2639} & \textbf{0.2748} \\
\midrule
pot \& penbag             & \underline{0.4733} & \textbf{0.4742} & 0.4548 & \underline{0.2380} & 0.2331 & \textbf{0.2451} \\
helmet \& headphone       & \textbf{0.4981} & 0.4626 & \underline{0.4964} & \underline{0.2762} & 0.2739 & \textbf{0.2765} \\
bucket \& doll            & \textbf{0.6135} & 0.5760 & \underline{0.6006} & 0.2600 & \underline{0.2560} & \textbf{0.2621} \\
robot toy \& dinosaur toy & 0.5346 & \underline{0.5513} & \textbf{0.5583} & 0.2821 & \textbf{0.2869} & \underline{0.2847} \\
cup \& cat figurine       & 0.4924 & \underline{0.5019} & \textbf{0.5427} & 0.2659 & \underline{0.2794} & \textbf{0.2803} \\

\bottomrule
\end{tabular}
\end{small}
\end{center}
\vskip -0.1in
\end{table}


\begin{table}[htb!]
\caption{Comparison of KID and LPIPS value between different parameter sets for optimization (Bold indicates the best value, underline represents the second-best value)}
\label{appendix-ablation-kid-lpips}
\vskip 0.1in
\begin{center}
\begin{small}
\begin{tabular}{lcccccc}
\toprule
\multirow{2}{*}{Target Concept} & \multicolumn{3}{c}{KID $\downarrow$} & \multicolumn{3}{c}{LPIPS $\uparrow$} \\
                                & $W_Q + W_K + W_V$ & $W_Q + W_V$ & $W_K + W_V$ & $W_Q + W_K + W_V$ & $W_Q + W_V$ & $W_K + W_V$ \\
\midrule

pot          & \textbf{0.1461} & \underline{0.1570} & 0.1695 & \textbf{0.6659} & 0.6517 & \underline{0.6578} \\
penbag       & \textbf{0.1675} & \underline{0.2091} & 0.2269 & \underline{0.6645} & 0.6616 & \textbf{0.6669} \\
helmet       & \textbf{0.1292} & 0.1355 & \underline{0.1305} & \underline{0.6649} & 0.6606 & \textbf{0.6651} \\
headphone    & \textbf{0.2018} & \underline{0.2094} & 0.2173 & 0.6089 & \textbf{0.6218} & \underline{0.6172} \\
bucket       & \underline{0.2208} & 0.2269 & \textbf{0.2171} & \underline{0.6874} & 0.6871 & \textbf{0.6968} \\
doll         & \underline{0.1477} & \textbf{0.1424} & 0.1515 & 0.6954 & \textbf{0.7100} & \underline{0.7006} \\
robot toy    & \underline{0.2982} & \textbf{0.2940} & 0.3108 & \underline{0.6768} & 0.6747 & \textbf{0.6802} \\
dinosaur toy & \underline{0.1720} & 0.1799 & \textbf{0.1554} & \underline{0.6087} & 0.6086 & \textbf{0.6130} \\
cup          & \textbf{0.0892} & 0.0988 & \underline{0.0895} & 0.6592 & \underline{0.6605} & \textbf{0.6805} \\
cat figurine & \textbf{0.4128} & 0.4508 & \underline{0.4239} & 0.6550 & \underline{0.6646} & \textbf{0.6767} \\
\midrule
pot \& penbag             & \textbf{0.1823} & \underline{0.1967} & 0.2259 & 0.6610 & \underline{0.6593} & \textbf{0.6672} \\
helmet \& headphone       & \underline{0.1120} & \textbf{0.1112} & 0.1174 & \textbf{0.6734} & \underline{0.6728} & 0.6662 \\
bucket \& doll            & \textbf{0.1068} & \underline{0.1090} & 0.1088 & 0.5637 & \underline{0.5738} & \textbf{0.5816} \\
robot toy \& dinosaur toy & \underline{0.2478} & 0.2459 & \textbf{0.2484} & \textbf{0.6271} & 0.6214 & \underline{0.6242} \\
cup \& cat figurine       & \underline{0.1075} & 0.1289 & \textbf{0.1039} & \textbf{0.6314} & 0.6244 & \underline{0.6268} \\

\bottomrule
\end{tabular}
\end{small}
\end{center}
\vskip -0.1in
\end{table}


\end{document}